\PassOptionsToPackage{table,svgnames,dvipsnames,prologue}{xcolor}
\documentclass[sigconf]{acmart}
\usepackage{balance}
\usepackage[ruled,vlined,linesnumbered,noline]{algorithm2e} % Use algorithm2e for styling
\usepackage{multirow}
\usepackage{makecell}
\SetKwInput{KwInput}{Input}   % Input keyword
\SetKwInput{KwOutput}{Output} % Output keyword

\definecolor{verylightgray}{gray}{0.9}
\AtBeginDocument{%
  }

\copyrightyear{2025}
\acmYear{2025}
\setcopyright{cc}
\setcctype{by}
\acmConference[MM '25]{Proceedings of the 33rd ACM International Conference on Multimedia}{October 27--31, 2025}{Dublin, Ireland}
\acmBooktitle{Proceedings of the 33rd ACM International Conference on Multimedia (MM '25), October 27--31, 2025, Dublin, Ireland}\acmDOI{10.1145/3746027.3755287}
\acmISBN{979-8-4007-2035-2/2025/10}

\begin{document}

\title{TolerantECG: A Foundation Model for Imperfect Electrocardiogram}


\author{Huynh Dang Nguyen}
\email{huynhnd11@fpt.com}
\affiliation{%
  \institution{AI Center, FPT Software}
  \city{Hanoi}
  \country{Vietnam}
}
\author{Trong-Thang Pham}
\email{tp030@uark.edu}
\affiliation{%
  \institution{EECS, University of Arkansas}
  \city{Fayetteville}
  \state{Arkansas}
  \country{USA}
}

\author{Ngan Le}
\email{thile@uark.edu}
\authornote{Both authors contributed equally to this research.}
\affiliation{%
  \institution{EECS, University of Arkansas}
  \city{Fayetteville}
  \state{Arkansas}
  \country{USA}
}
% \affiliation{%
%   \institution{FPT Software AI Center}
%   \country{Vietnam}
% }

\author{Van Nguyen}
\email{vannth19@fpt.com}
\authornotemark[1]
\affiliation{%
  \institution{AI Center, FPT Software}
  \city{Hanoi}
  \country{Vietnam}
}
\renewcommand{\shortauthors}{Huynh Dang Nguyen, Trong-Thang Pham, Ngan Le, and Van Nguyen}
\newcommand{\quotes}[1]{``#1''}

\begin{abstract}
  The electrocardiogram (ECG) is an essential and effective tool for diagnosing heart diseases. However, its effectiveness can be compromised by noise or unavailability of one or more leads of the standard 12-lead recordings, resulting in diagnostic errors or uncertainty. To address these challenges, we propose \emph{TolerantECG}, a foundation model for ECG signals that is robust to noise and capable of functioning with arbitrary subsets of the standard 12-lead ECG. TolerantECG training combines contrastive and self-supervised learning frameworks to jointly learn ECG signal representations alongside their corresponding knowledge-retrieval-based text report descriptions and corrupted or lead-missing signals. Comprehensive benchmarking results demonstrate that TolerantECG consistently ranks as the best or second-best performer across various ECG signal conditions and class levels in the PTB-XL dataset, and achieves the highest performance on the MIT-BIH Arrhythmia Database. The source is available at this link: \href{https://github.com/Fsoft-AIC/TolerantECG}{https://github.com/Fsoft-AIC/TolerantECG}
\end{abstract}




% \received{10 April 2025}

\maketitle

\section{Introduction}

The electrocardiogram (ECG) is a non-invasive and widely used technique for monitoring and diagnosing cardiac health. By recording the heart's electrical activity, ECG provides critical insights into rhythm disturbances, myocardial ischemia, and other cardiac abnormalities. However, the reliability of ECG analysis often hinges on the quality and completeness of the recorded signals. In real-world scenarios, ECG signals can easily become corrupted due to body movements, poor electrode contact, or even respiration \cite{ecg-noises}. Furthermore, not all ECG measuring devices capture all 12 leads. Daily used devices, such as smartwatches and Holter monitors, typically provide data from only a few leads, limiting the comprehensiveness of the analysis \cite{smartwatch_beyond_afib, smartwatch, smartwatch2, 1-lead}. 

In addition to hardware limitations, artifacts can significantly impact the performance of cardiac diagnostics \cite{noise-impact, noise-impact2, noise-impact3}. To address this issue, numerous methods for ECG denoising have been developed, ranging from traditional signal filtering techniques to machine learning and deep learning approaches \cite{denoise-dwt, denoise-transformer, denoise-mamba}. However, prepending such a denoising module to an ECG interpretation algorithm can introduce delays and increase computational complexity. Furthermore, it does not address the lead-missing problem.

In this paper, we introduce TolerantECG, a framework designed to learn robust ECG representations even in the presence of missing leads, noise artifacts, or both. Our pretraining framework combines signal-report contrastive learning and self-supervised learning with a dual-mode distillation mechanism that alternates between addressing different imperfect ECG conditions. The model leverages multimodal contrastive learning to refine ECG signal representations through the guidance of detailed report descriptions. Unlike \cite{esi, esi-rag}, which employ a ChatGPT-based Retrieval-Augmented Generation (RAG) system, we employs a Large Language Model (LLM)-free knowledge retrieval framework called Cardiac Feature Retrieval (CFR) to enhance ECG report descriptions. Using this retrieval module, CFR is able to provide relevant diagnostic criteria from a public ECG feature database and combines them with patient characteristics to generate comprehensive and descriptive ECG reports. Additionally, we present a self-supervised learning approach featuring an alternating training pipeline, enabling TolerantECG to iteratively and efficiently learn from various ECG conditions. Our main contributions can be summarized as follows:

\begin{itemize}
    \item We propose TolerantECG, an ECG representation learning framework that combines contrastive learning with associated detailed text reports and self-supervised learning with dual-mode distillation under different ECG signal conditions. The resulting model produces ECG representations that tolerate different types and levels of imperfect conditions, specifically noise corruption, lead-missing, or both.
    \item We develop Cardiac Feature Retrieval, a simple yet effective knowledge retrieval system designed to retrieve detailed waveform characteristics associated with each cardiac condition.
    \item We conduct benchmarking across all classification levels of the PTB-XL and MIT-BIH datasets under various imperfect conditions, facilitating a comprehensive performance analysis that assesses both the transferability of pretrained models to diverse downstream tasks and their robustness to corrupted signals.
\end{itemize}

\section{Related Works}
\subsection{Representation learning}
The recent advancements in deep learning have been significantly influenced by self-supervised and contrastive learning methods. Among these, CLIP (Contrastive Language-Image Pre-training) \cite{clip} and DINO (Self-Supervised Vision Transformers) \cite{dino} stand out as pioneering frameworks exploring distinct approaches to vision representation learning. 

CLIP revolutionized the paradigm of vision representation learning by leveraging natural language supervision instead of traditional labeled datasets. It employs a contrastive objective to align images and text, enabling the model to predict which text snippet pairs with a given image in a batch of examples. With a dataset of 400 million (image, text) pairs, CLIP demonstrated strong zero-shot transfer performance across more than 30 tasks, achieving results competitive with supervised baselines on datasets such as ImageNet. The ability to directly use natural language descriptions for image classification underscores its versatility, particularly for tasks with limited labeled data.

% Self-Supervised Vision Transformers
DINO introduced a self-supervised learning approach tailored to Vision Transformers (ViTs), achieving remarkable results by leveraging features that emerge naturally during training. Unlike CLIP, which relies on paired image-text data, DINO employs self-distillation without labels, where a student network learns to mimic a teacher network updated via momentum encoding. This framework showed that self-supervised ViTs can produce features explicitly encoding semantic segmentation and outperform prior self-supervised methods in linear evaluations on ImageNet. Notably, DINO demonstrates the power of combining ViTs with techniques such as multi-crop training and careful architectural choices like smaller patch sizes.

% Comparisons and Emerging Properties
While both methods highlight the potential of self-supervised and contrastive learning, their methodologies differ fundamentally. CLIP's reliance on large-scale paired data provides generalizability and flexibility in zero-shot scenarios, whereas DINO's approach emphasizes emergent semantic understanding and feature quality without requiring explicit labels. Additionally, although both are proposed for computer vision, they provide the fundamental framework to inspire similar developments in other modalities.

% Both frameworks underline the ongoing shift in computer vision towards reducing dependence on labeled datasets while enhancing model generalizability and task adaptability.

\subsection{Self-supervised learning from 12-lead ECG}

Self-supervised learning is a method that enables models to learn rich and detailed representations of input data by using the data itself to generate training signals, rather than depending on manually labeled annotations. This technique is particularly advantageous for scenarios where datasets are scarce or lack comprehensive labels. Unlike supervised learning, which requires a large volume of annotated examples, self-supervised learning automatically derives supervisory signals from the structure of the data, enabling the model to learn useful feature representations without human intervention. This is especially crucial in biomedical domains such as ECG, where high-quality annotations are not only limited but also expensive to obtain. Annotating ECG signals requires a team of expert physicians to meticulously examine each waveform, identify abnormalities, and provide accurate diagnoses, making large-scale manual labeling impractical and time-consuming. As a result, leveraging self-supervised learning can significantly reduce the reliance on expert-labeled data while still achieving strong performance in downstream tasks such as arrhythmia classification.

\citet{self-supervised_ecg} explore three distinct self-supervised learning approaches: SimCLR \cite{simclr}, BYOL \cite{byol}, and CPC \cite{cpc}. These models are trained on a dataset comprising 54,566 ECG samples, each sampled at 100 Hz, while incorporating a range of signal transformation pairs to enhance representation learning. Transformations include techniques such as segment signal cropping, time period masking, and noise injection, which help models learn invariant features and generalize better to real-world variations in ECG signals. By contrasting differently augmented versions of the same signal, the models are encouraged to focus on essential features while discarding irrelevant variations. These experiments demonstrate the effectiveness of self-supervised learning in extracting meaningful representations from ECG signals, paving the way for improved diagnostic performance, transferability to new datasets, and robust feature learning in clinical applications.

\subsection{ECG foundation models}
Foundation models represent a transformative approach across domains including the field of electrocardiogram analysis, addressing challenges such as dependency on large-scale labeled datasets and improving downstream classification performance. Foundation models leverage transfer learning to provide a unified representation that can be adapted to various downstream tasks with minimal additional labeled data. The ECG-FM model \cite{ecg-fm} adopts a transformer-based architecture pretrained on over 2.5 million ECG samples using ECG-specific augmentations and a combination of contrastive learning and signal masking objectives. This comprehensive approach enables ECG-FM to learn rich contextual embeddings, improving its generalization across tasks such as prediction of select conditions from ECG interpretations, reduced left ventricular ejection fraction, and abnormal cardiac troponin (cTn).

MERL \cite{merl} employs a dual-alignment strategy to learn ECG representations directly from both ECG signals and their corresponding text reports. This approach integrates inter-modality alignment, which captures relationships between the ECG and text domains, and intra-modality alignment, which refines ECG representations. By leveraging these two alignment mechanisms, MERL is able to generate high-quality ECG embeddings that facilitate zero-shot cardiac classification, enabling the identification of previously unseen diseases without requiring additional labeled data.

Similarly, the ESI model \cite{esi} employs a multimodal pretraining framework that learns jointly from ECG signals and their corresponding textual descriptions. By incorporating a retrieval-augmented generation pipeline, ESI aligns ECG waveforms with enhanced textual semantics, enabling robust representation learning. This multimodal approach demonstrates improved performance on downstream tasks such as arrhythmia classification and ECG-based subject identification, showcasing its ability to bridge waveform and semantic content effectively.

\section{Method}
\subsection{Cardiac Feature Retrieval (CFR)}

\begin{figure*}[h]
  \centering
  \includegraphics[width=0.9\textwidth]{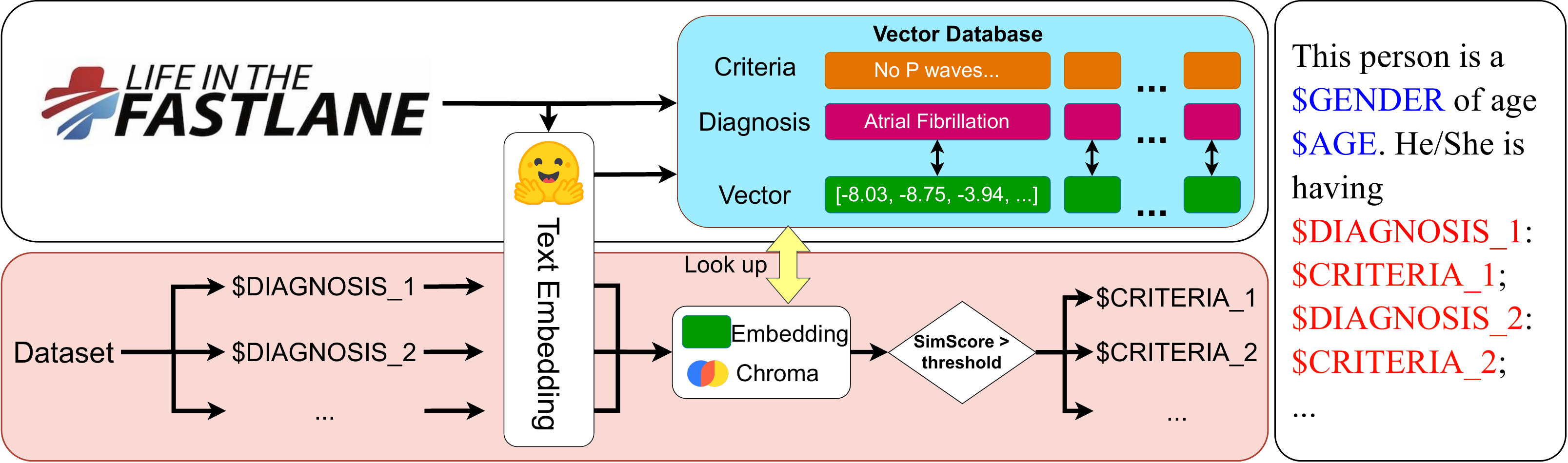}
  \caption{The Cardiac Feature Retrieval (CFR) pipeline employs information gathered from \textit{Life in the Fastlane} website to retrieve relevant waveform criteria for each cardiac disease. Diagnoses are encoded into vector embeddings and stored in a vector database.}
  \label{fig:retrieval}
\end{figure*}

\citet{esi} and \citet{esi-rag} employs a ChatGPT-based Retrieval-Augmented Generatio (RAG) system to extract waveform criteria of various cardiac diseases from provided textbooks as external knowledge. However, relying on OpenAI's API for text embeddings and tokens can become costly for large datasets, and ChatGPT’s reliability, particularly in the medical domain, is not guaranteed \cite{shen2023chatgpttrustmeasuringcharacterizing, Cappellani2024}. To address these challenges, we develop Cardiac Feature Retrieval (CFR), a knowledge retrieval system that can directly retrieve text description of waveform features associated with a given diagnosis without the need for a Large Language Model (LLM). The retrieved waveform features, along with given metadata from ECG recordings including patient's gender and age, are used to construct a detailed report description for each ECG signal. The constructed text description is used during contrastive learning to enhance the semantic representations of ECG signals. The schematic of the CFR method is depicted in Figure \ref{fig:retrieval}.

\subsubsection{Diagnostic-Description Database Construction}

In order for the retrieval system to work, we first construct a dictionary database where the keys represent cardiac diseases and the values describe the associated waveform features. For this phase, we scrape relevant information from \textit{LITFL} \cite{LITFL}. The keys are then encoded using the \textit{all-MiniLM-L6-v2} model \cite{sentence-bert}, and the resulting embeddings are stored in a vector database for use in the retrieval system. To effectively organize the database, same as \cite{esi, esi-rag}, we use the \textit{Chroma} management tool, which offers robust functionality and integrates seamlessly with the \textit{LangChain} Python library.

\subsubsection{Detailed Report Construction}

ECG diagnostic reports typically focus on information related to heart conditions, such as standard clinical labels, SCP (Standard Communications Protocol for Computer-Assisted ECG) statements, diagnostic interpretations, and machine-generated findings. However, these reports often lack explicit details about the ECG waveform features associated with each condition. Moreover, directly using diagnostic reports poses challenges for contrastive learning, as textual descriptions can exhibit high similarity, making it difficult to effectively distinguish between different cases \cite{ecg-chat}. To overcome this limitation, we use CFR to retrieve relevant cardiac waveform criteria for every heart condition as additional information for the model.

\begin{figure}[ht]
    \begin{center}
        \centerline{\includegraphics[width=0.9\columnwidth]{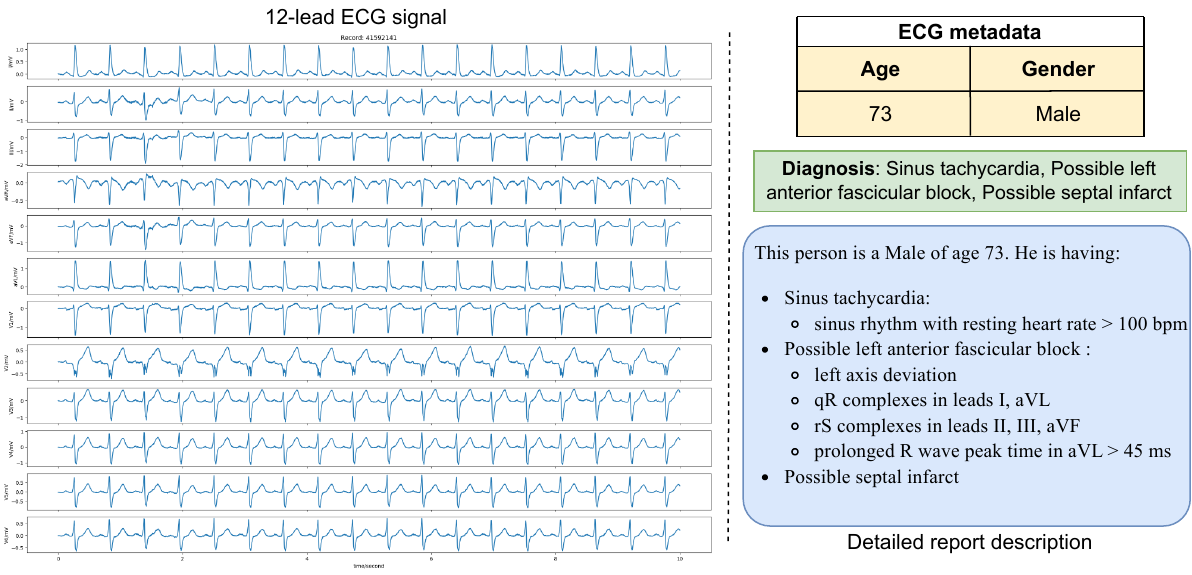}}
        \caption{Example of a 12-lead ECG recording with its associated metadata and diagnoses, and the detailed report description constructed by our method.}
        \label{fig:cfr-example}
    \end{center}
\end{figure}

Using the CFR pipeline, we can generate a comprehensive report description for each ECG signal based on the corresponding SCP statements. Since a single ECG signal may represent multiple diagnoses, often abbreviated with acronyms, we process them individually by expanding the abbreviations (e.g., converting \quotes{PAC} to \quotes{Premature Atrial Complex}). Each expanded phrase is then converted into a text embedding using the \textit{all-MiniLM-L6-v2} text embedding model. Cosine similarities between the input embedding and each key are calculated, and the key with the highest similarity score that is higher than a predefined threshold is identified. The waveform criteria associated with the selected key are then used to construct the final detailed report. Figure \ref{fig:cfr-example} shows an example with an ECG recording, its associated metadata and diagnoses, and the detailed report constructed by CFR.

\subsection{TolerantECG framework}
\label{sec:main-framework}
\begin{figure*}[ht]
    %\vskip 0.5in
    \begin{center}
        \centerline{\includegraphics[width=0.9\textwidth]{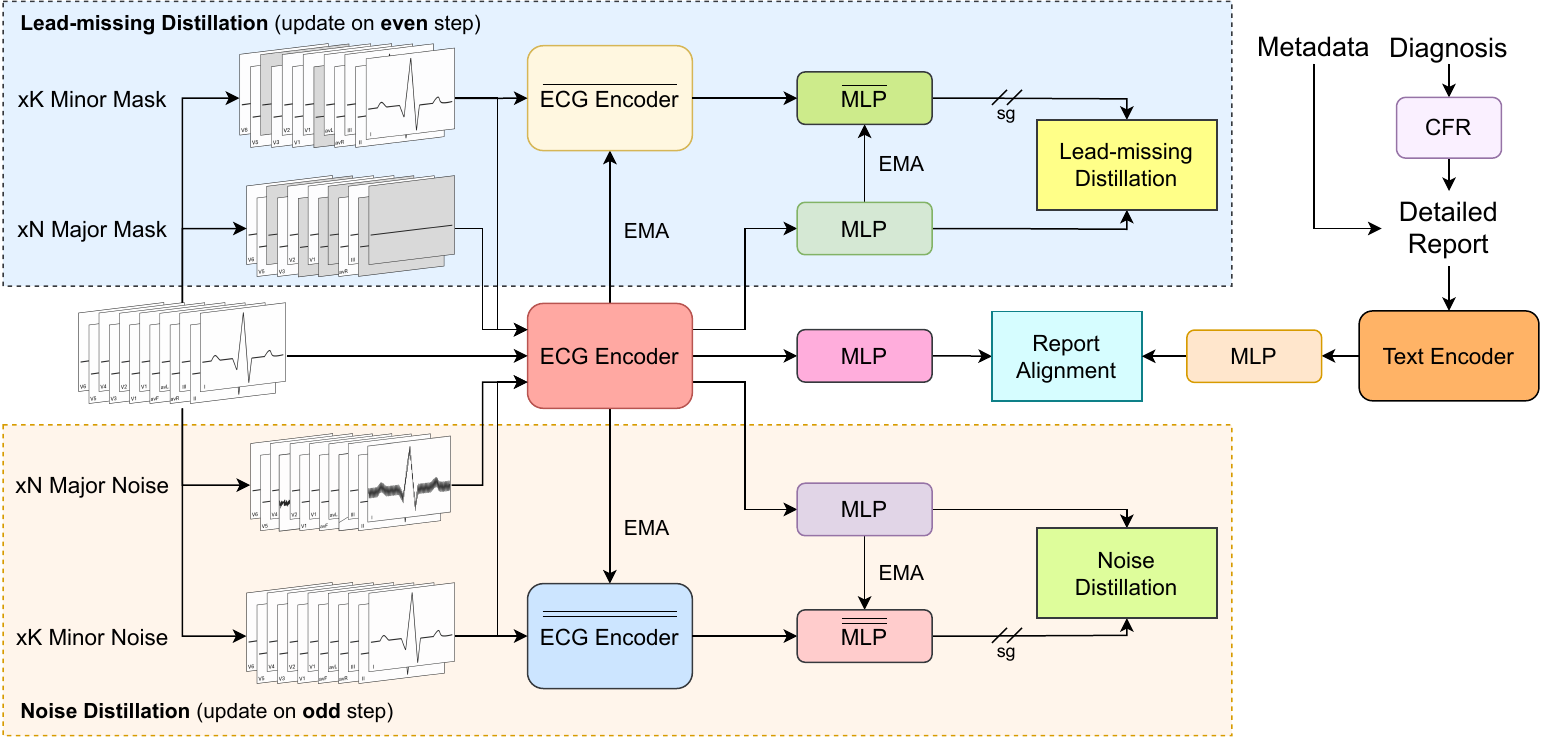}}
        \caption{TolerantECG training framework, which comprises (1) a Report Alignment module that aligns ECG signals with detailed text reports and (2) a self-supervised learning module with lead-missing distillation and noise distillation (DuoDistill). In DuoDistill, the student model (\textit{ECG Encoder}) and two teacher models ($\overline{\textit{ECG Encoder}}$ and $\overline{\overline{\textit{ECG Encoder}}}$) have the same architecture, and the distillation steps occur in a ping-pong manner, alternating between the two different distillation tasks. 
        The student processes both Minor and Major augmentations, where Minor augmentation retains a large portion of the original signal while Major augmentation preserves a small portion. These augmentations involve lead masking and noise addition.
        A stop-gradient (sg) operator is applied to the teachers to propagate gradients only through the student. The total loss is the combination of the contrastive learning loss and the self-supervised loss.
        The \textit{ECG Encoder} utilizes randomly initialized ConvNeXt V2, while pretrained BioLinkBERT is used as the \textit{Text Encoder}.}
        %The outputs from all encoders are fed into a task-specific Multi-Layer Perceptron (MLP) layer, which projects the representations into same dimension before computing the loss functions.
        \label{fig:arch} 
    \end{center}    
    \vspace{-0.5cm}
\end{figure*}

Our proposed method, TolerantECG, consists of two main components:
\begin{itemize}
    % Contrastive Learning
    \item \textbf{Report Alignment (\textit{ReportAlign})}: This module learns ECG representations that are aligned with their detailed textual reports, enforcing the semantic alignment of the two modalities.
    % Self-supervised learning
    \item \textbf{Self-supervised learning with dual-mode distillation (\textit{DuoDistill})}: To enable the model to handle imperfect ECG signals, we propose a self-supervised learning approach that alternates between different imperfect signal conditions, specifically ECG recordings with missing leads and those corrupted by different types and levels of noise. This enhances robust performance across imperfect scenarios.
\end{itemize}

\subsubsection{Report Alignment (ReportAlign)}
TolerantECG employs a dual-modal contrastive learning approach that integrates ECG signals with their corresponding detailed report descriptions. By minimizing the distance between positive signal-text pairs and maximizing the separation from negative pairs, it enhances the model's ability to capture meaningful correlations across modalities.

To process the ECG signal, we adopt a modified one-dimensional ConvNeXt V2 architecture \cite{convnextv2} as the \textit{ECG encoder}. By replacing the 2D convolutional layers with 1D convolutional layers using the same arguments, this encoder effectively captures features from the ECG waveforms. The output from the final layer is passed through an average pooling operation along the length dimension to generate the ECG embedding.

In addition, to extract meaningful features from the report description, we employ BioLinkBERT \cite{biolinkbert} as the \textit{Text encoder}. BioLinkBERT is a BERT-based model \cite{bert} pretrained on extensive biomedical and clinical text datasets from PubMed. The CLS token from the final layer serves as the embedding for the entire text input sequence. 
We set random initial weights for the \textit{ECG encoder} and initialize the \textit{Text encoder} with the pretrained weights. During pretraining, the Text encoder is kept frozen \cite{frozen-llm, merl, esi, ecg-chat}.

After extracting features from both modalities—ECG signals and textual report descriptions—TolerantECG contrasts the two embeddings using the CLIP framework \cite{clip} with the contrastive loss expressed as:

\resizebox{\linewidth}{!}{
$
    \mathcal{L}_{\text{ReportAlign}}(\mathbf{S}, \mathbf{T}) = -\frac{1}{2} 
    \Bigg(
        \underbrace{
            \sum_{i=1}^{B} \log \frac{\exp(\mathbf{S}_i^\top \mathbf{T}_i / \tau)}
            {\sum_{j=1}^{B} \exp(\mathbf{S}_i^\top \mathbf{T}_j / \tau)}
        }_{\text{ecg-to-text}}
        + 
        \underbrace{
            \sum_{i=1}^{B} \log \frac{\exp(\mathbf{T}_i^\top \mathbf{S}_i / \tau)}
            {\sum_{j=1}^{B} \exp(\mathbf{T}_i^\top \mathbf{S}_j / \tau)}
        }_{\text{text-to-ecg}}
    \Bigg)
$
}

where $\mathbf{S}_i$ and $\mathbf{T}_i$ are the projected embeddings from the \textit{ECG encoder} and \textit{Text encoder} for the $i$-th signal-text pair, $B$ is the training batch size, and $\tau$ is the logit temperature scaling factor.

\subsubsection{Self-supervised learning with dual-mode distillation (DuoDistill)}

%To enable TolerantECG to handle different imperfect signals, such as those with missing leads or corrupted by noise, we develop a self-supervised learning pipeline tailored for each scenario, leveraging the self-distillation with no labels (DINO) framework \cite{dino}. 

To enable handling of different imperfect signals, such as those with missing leads or corrupted by noise, we develop a self-supervised learning pipeline tailored for each scenario, leveraging the self-distillation with no labels (DINO) framework \cite{dino}. 

DINO is designed for image representation learning using a multi-view learning approach, which aligns representations generated by the student and teacher models. The student processes both local views—partial patches of the input image—and global views, where a more extensive portion of the input is visible. In contrast, the teacher model operates exclusively on the global views. The training objective ensures that the student learns to generate representations that closely resemble the teacher’s global view representations, fostering consistency across different perspectives of the input.

Inspired by this framework, we impose different levels of imperfect signal conditions with the teacher seeing the versions that are more similar to the original/perfect signals and the student seeing less similar versions. Unlike DINO, we employ a dual-teacher-single-student framework where the student (i.e. \textit{ECG encoder}) learns from two teachers: one is specialized in handling lead-missing condition and the other is specialized in handling noisy condition. In other words, we conduct two distillation processes in the same training framework: lead-missing distillation and noise distillation, where each teacher guides the student to learn meaningful and consistent representations regardless of how corrupted the data are. The objective is to enable the student model to handle various imperfect signal scenarios while maintaining performance comparable to ideal signal conditions. 

Specifically, for both distillation modules, the input signal is corrupted multiple times with $N$ variations, referred to as Major Mask/Noise. In contrast, the $K$ other versions, where the majority of the original signal is preserved, are called Minor Mask/Noise. In this paper, we set $N = 8$ and $K = 2$. 

For lead-missing distillation, the ECG is masked at two levels: $N$ Major Mask (keeping 1 to 6 leads) and $K$ Minor Mask (keeping 6 to 12 leads), with the remaining leads zeroed out. The student model processes all $N + K$ augmented signals as input, while the teacher only receives the $K$ Minor Mask signals.

For noise distillation, we introduce noise to the ECG signal using the MIT-BIH Noise Stress Test Database \cite{mit-bih} as the noise source. This dataset contains three common types of noise found in ambulatory ECG recordings: baseline wander, muscle artifact, and electrode motion artifact. Each noise type provides a total of 1 hour of recordings sampled at 360 Hz.

To prepare noise, we first resample it to a 500 Hz sample rate to match the ECG signals, then randomly extract 10-second segment. Each type of noise is applied to the ECG signal with a probability of $0.7$. To enhance data diversity, the combined noise is added to each lead of the original signal with a probability of $0.5$. The Signal-to-Noise Ratio (SNR) for noise injection is uniformly sampled between $-10$~dB and $0$~dB, ensuring a diverse range of noise levels. This process creates $N$ Major Noise signals. Since $K=2$, we use the original signal and the signal filtered by high-pass and low-pass filters as Minor Noise. If $K>2$, additional Minor Noise signals are generated by adding noise at high SNR values. The student processes all $N$ Major Noise signals and $K$ Minor Noise signals, while the teacher only receives the Minor Noise signals.

For the self-supervised phase, we adopt the DINO loss function to guide the training process:

\resizebox{\linewidth}{!}{
$
    \mathcal{L}_{\text{DuoDistill}}(\mathbf{q}, \mathbf{p}) = -\mathbb{E} \Bigg[ 
    \sum_{i=1}^K \sum_{j=1\text{, }j \neq i}^{K + N} \sum_{d=1}^D
    \text{softmax} \left( \frac{\mathbf{q}^i_d - \mathbf{c}_d}{\tau_t} \right)
    \cdot 
    \log \Bigg( 
    \text{softmax} \left( \frac{\mathbf{p}^j_d}{\tau_s} \right) 
    \Bigg) \Bigg]
$
}

\noindent where \( \mathbf{q}^i_d \) and \( \mathbf{p}^j_d \) are the teacher and student logits at index $d$, while \( \tau_t \) and \( \tau_s \) are their respective temperatures. The centering term for the teacher's logits is \( \mathbf{c}_d \), and $K$, $N$ and $D$ denote the number of Minor Mask/Noise, Major Mask/Noise, and the logit dimension, respectively.

Finally, the total loss used for gradient calculation and backpropagation is formulated as follows:
\begin{equation*}
    \mathcal{L}_{\text{Total}} = \alpha \mathcal{L}_{\text{ReportAlign}}(\mathbf{S}, \mathbf{T}) + \beta \mathcal{L}_{\text{DuoDistill}}(\mathbf{q}, \mathbf{p}), 
\end{equation*}
where $\alpha$ and $\beta$  are hyper-parameters specifying the weights for each component. For this experiment, we assign equal weights to the two loss components, setting $\alpha = \beta = 1$.

\subsubsection{Alternating training}
Since the teachers’ weights are updated using an exponential moving average of the student’s weights after each step \cite{dino}, standard backpropagation would lead to identical teacher models. To address this issue, we introduce an alternating teacher-updating pipeline. The contrastive learning pipeline remains consistent across all steps.

During the pre-training phase, at every even step, the original ECG signals in the current batch are masked at varying levels for both the student and the teacher. Once the loss for the masked branch is computed, it is combined with the contrastive loss, and gradient descent is applied. The lead-missing expert teacher is then updated using the updated weights of the \textit{ECG encoder}.

At the odd steps, the ECG signals are corrupted with up to all three types of noise. The process follows the same flow as the lead-missing branch; however, in this case, the noise expert teacher is updated after backpropagation. This alternating strategy resembles a student progressively learning from one subject to another. 
The entire training procedure is outlined in Algorithm \ref{alg:training}.

\begin{algorithm}[t]
\caption{Training process of TolerantECG}
\label{alg:training}
\KwInput{
    ECG signal $E$, report description $R$; 
    mask teacher $\overline{\textit{ECG\_Encoder}}$, noisy teacher $\overline{\overline{\textit{ECG\_Encoder}}}$, 
    student model \textit{ECG\_Encoder}, 
    text model \textit{Text\_Encoder}, 
    total steps $S$
}

\For{$s = 0$ to $S-1$}{
    % ---------------------------
    % Step 1: Cross-modal alignment
    % ---------------------------
    $\mathbf{S} = \textit{ECG\_Encoder}(E)$ \tcp*[f]{ECG embedding}\\
    $\mathbf{T} = \textit{Text\_Encoder}(R)$ \tcp*[f]{Text embedding}\\

    % ---------------------------
    % Step 2: Self-supervised DuoDistill
    % ---------------------------
    \eIf{$s$ is even}{
        $(E_\text{major}, E_\text{minor}) = \text{add\_mask}(E)$ \\
        $\mathbf{q} = \overline{\textit{ECG\_Encoder}}(E_\text{minor})$ \\
        $\mathbf{p} = \textit{ECG\_Encoder}([E_\text{minor}, E_\text{major}])$
    }{
        $(E_\text{major}, E_\text{minor}) = \text{add\_noise}(E)$ \\
        $\mathbf{q}= \overline{\overline{\textit{ECG\_Encoder}}}(E_\text{minor})$ \\
        $\mathbf{p} = \textit{ECG\_Encoder}([E_\text{minor}, E_\text{major}])$
    }

    % ---------------------------
    % Step 3: Optimization
    % ---------------------------
    $\mathcal{L}_{\text{Total}} = \alpha \mathcal{L}_{\text{ReportAlign}}(\mathbf{S}, \mathbf{T}) + \beta \mathcal{L}_{\text{DuoDistill}}(\mathbf{q}, \mathbf{p})$ \\
    
    Backpropagate $\mathcal{L_{\text{Total}}}$ \\

    \eIf{$s$ is even}{
        update($\overline{\textit{ECG\_Encoder}} \leftarrow \textit{ECG\_Encoder}$)
    }{
        update($\overline{\overline{\textit{ECG\_Encoder}}} \leftarrow \textit{ECG\_Encoder}$)
    }
}
\end{algorithm}

\section{Experiments}
\subsection{Pretraining}
\label{sec:pretrain}

We use the MIMIC-IV-ECG dataset \cite{mimic}, provided by PhysioNet \cite{physionet}, as the pretraining corpus. It comprises approximately 800,000 12-lead diagnostic electrocardiograms from nearly 160,000 unique patients. Each recording is 10 seconds long and sampled at 500 Hz, accompanied by demographic information and machine-generated ECG reports. For data quality, we exclude samples that are entirely zero or contain NaN values.

The pretraining method follows the framework described in Section~\ref{sec:main-framework}. We use AdamW optimizer with a learning rate of $3 \times 10^{-5}$, for a total of just 10 epochs on 4 Nvidia A100 80 GB GPUs. Each GPU takes a batch size of 32.

\subsection{Downstream Evaluation}

For transfer learning to each downstream task, we attach a simple linear classifier head to the ECG encoder. The ECG encoder is initiated with its pretrained weights while the downstream head is randomly initialized, and the whole network is finetuned.

We use the PTB-XL dataset \cite{ptb-xl}, a clinical 12-lead ECG waveform dataset containing 21,837 records from 18,885 patients, each is 10 seconds long. The ECG waveforms were professionally annotated as a multi-label dataset. It encompasses a wide range of diagnostic categories, including a significant proportion of healthy recordings. The dataset is enriched with additional metadata, such as demographic details, supplemental diagnostic statements, diagnosis likelihoods, and manually annotated signal properties. We follow the train-validation-test splitting guide in \cite{ptb-xl}.

Additionally, to further evaluate our method in the lead-missing scenario, we incorporate the MIT-BIH Arrhythmia Database \cite{mit-bih-data}, which contains only two leads selected from lead II, V1, V2, V4, and V5. This dataset consists of 48 half-hour ECG recordings sampled at 360 Hz. We segment each recording into multiple 10-second clips and resample them to 500 Hz to maintain consistency with other dataset. Following the approach of \cite{merl}, we use the annotated beats within each segment as the corresponding labels, resulting in a multi-label classification dataset with five classes: Normal, LBBB, RBBB, PAC, and PVC. The dataset is split into training, validation, and test sets with a ratio of 70\%, 10\%, and 20\%, respectively.

% Additionally, we incorporate the CPSC2018 dataset, which consists of 6,877 12-lead ECG recordings with durations ranging from 6 to 144 seconds, sampled at 500 Hz \cite{cpsc2018}. Since the original source is no longer publicly accessible, we utilize the waveform data provided by the PhysioNet 2021 Challenge \cite{physionet} and adopt the labels and data splits from the MERL \cite{merl} implementation. For ECG signals shorter than 10 seconds, we pad them with zeros; for those longer than 10 seconds, we truncate them to exactly 10 seconds.

Because, to the best of our knowledge, there is no publicly available imperfect ECG dataset, we construct our own using the PTB-XL \cite{ptb-xl} datasets. For each finetuning epoch, besides the original set, we form another three datasets by performing data augmentation to represent three imperfect scenarios: lead-missing, noisy, and a combination of lead-missing and noisy. We then aggregate all four datasets into a single comprehensive dataset that covers various ECG conditions. Below is the detailed description of each data category in this dataset. The validation and test sets are constructed in the same manner but remain fixed. Moreover, for the MIT-BIH Arrhythmia Database, we only combine it with another noisy version.

\begin{itemize}
    \item \textit{Original}: The original 12-lead PTB-XL dataset.
    \item \textit{Lead-missing}: Following \citet{ecg-fm}, this set is formed by randomly zero out (i.e. masking) each lead in the \textit{Original} set with a probability $p_{mask}=0.5$ but with at least one lead retained.
    \item \textit{Noisy}: This set is formed by adding noise to one or more leads in the \textit{Original} test set with a probability $p_{noise} = 0.5$. The added noise is a combination of up to three noise types from the MIT-BIH Noise Stress Test Database \cite{mit-bih}, with each noise type having a 0.7 probability of being applied. The SNR is uniformly sampled between -10~dB and 0~dB.
    \item \textit{Lead-missing \& Noisy}: This set is created by randomly applying both lead masking and noise corruption to the \textit{Original} dataset. First, the \textit{Original} ECG recording is corrupted with a combination of noises, as described earlier. Then, the noisy ECG is further processed by applying lead masking, where each lead is zero-ed with a probability of $p_{mask}$.
\end{itemize}

\subsection{Results}
\subsubsection{Downstream performance}

\begin{table*}[ht]
    \caption{\textbf{AP} and \textbf{AUC} score on various PTB-XL's class levels (i.e. downstream tasks). 
    The \textbf{bold} values represent the best score, while the \underline{underlined} values denote the second best.
    }
    \centering\scalebox{0.9}{
    \begin{tabular}{c | c | c c | c c | c c | c c | c c | c c}
        \hline
         \multirow{2}{*}{\textbf{Augmentation}} & 
         \multirow{2}{*}{\textbf{Model}} & 
         \multicolumn{2}{c|}{\textbf{Super-Diag}} & 
         \multicolumn{2}{c|}{\textbf{Sub-Diag}} & 
         \multicolumn{2}{c|}{\textbf{Diag}} & 
         \multicolumn{2}{c|}{\textbf{Rhythm}} & 
         \multicolumn{2}{c|}{\textbf{Form}} & 
         \multicolumn{2}{c}{\textbf{All}} \\
         % \cline{3-14}
         & & AP & AUC & AP & AUC & AP & AUC & AP & AUC & AP & AUC & AP & AUC \\
         \Xhline{4\arrayrulewidth}
         
         \multirow{7}{*}{Original} 
         & SimCLR & .744 & .893 & .359 & .879 & .239 & .896 & .490 & .952 & .163 & .812 & .246	& .876 \\
         % \cline{2-8}
         & BYOL & .752 & .898 & .408	& .911 & .276 & \underline{.914} & .492	& .953 & .197 & .841 & .272 & \underline{.895} \\
         % \cline{2-8}
         & CPC & .772 & .904	& .312 & .831 & .142 & .792	& .369 & .909 & .170 & .791 & .165 & .817 \\
         % \cline{2-8}
         & METS & \underline{.812} & \underline{.924} & .403 & .898 & .242 & .880 & .414	& .884 & .149 & .810 & .210	& .847 \\
         % \cline{2-8}
         & MERL & .774 & .907 & .334 & .845 & .186 & .849 & .330 & .844 & .121 & .787 & .165 & .802 \\
         % \cline{2-8}
         & ECG-FM & .792 & .917 & \underline{.485} & \underline{.917} & \underline{.350} & .911 & \textbf{.601} & \textbf{.967} & \textbf{.241} & \underline{.868} & \underline{.319} & .874 \\
         % \cline{2-8}
         \rowcolor{verylightgray}
         & $\textbf{TolerantECG}_{\textbf{(ours)}}$ & \textbf{.814} & \textbf{.926} & \textbf{.499} & \textbf{.925} & \textbf{.360} & \textbf{.934} & \underline{.530} & \underline{.958} & \underline{.234} & \textbf{.884} & \textbf{.340} & \textbf{.915} \\
         \Xhline{4\arrayrulewidth}

         \multirow{7}{*}{\parbox{1.2cm}{\centering Lead-missing}}
         & SimCLR & .699 & .869 & .326 & .858 & .206 & .877 & .489 & .951 & .148& .797 & .225 & .860 \\
         % \cline{2-8}
         & BYOL & .710 & .876 & .367	& .893 & .237 & .888 & .493 & .950 & .180 & .813 & .252 & \underline{.877} \\
         % \cline{2-8}
         & CPC & .708 & .874 & .262 & .798 & .119 & .775 & .361 & .887 & .137 & .763 & .148 & .801 \\
         % \cline{2-8}
         & METS & .740 & .889 & .337 & .862 & .197 & .843 & .372 & .856 & .116 & .765 & .176 & .805 \\
         % \cline{2-8}
         & MERL & .708 & .871 & .278 & .812 & .152 & .815 & .281 & .814 & .092 & .726 & .138 & .766 \\
         % \cline{2-8}
         & ECG-FM & \textbf{.775} & \textbf{.907} & \textbf{.460} & \textbf{.907} & \textbf{.322} & \underline{.903} & \textbf{.585} & \textbf{.967} & \textbf{.223} & \underline{.863} & \underline{.302} & .870 \\
         % \cline{2-8}
         \rowcolor{verylightgray}
         & $\textbf{TolerantECG}_{\textbf{(ours)}}$ & \underline{.773} & \underline{.905} & \underline{.455} & \underline{.906} & \underline{.315} & \textbf{.922} & \underline{.529} & \underline{.960} & \underline{.208} & \textbf{.870} & \textbf{.310} & \textbf{.903} \\
         \Xhline{4\arrayrulewidth}

         \multirow{7}{*}{Noisy} 
         & SimCLR & .730 & .887 & .344 & .873 & .228 & .891 & .497 & .948 & .140 & .800 & .240 & .871 \\
         % \cline{2-8}
         & BYOL & .739 & .889 & .400 & .905 & .262 & \underline{.905} & .478 & .949 & .159 & .811 & .265 & \underline{.883} \\
         % \cline{2-8}
         & CPC & .742 & .888 & .283 & .819 & .113 & .757 & .354 & .881 & .133 & .772 & .155 & .810 \\
         % \cline{2-8}
         & METS & \underline{.784} & \underline{.911} & .375 & .863 & .215 & .849 & .359 & .873 & .145 & .775 & .183 & .823 \\
         % \cline{2-8}
         & ECG-FM & .770 & .905 & \underline{.460} & \underline{.911} & \underline{.344} & .900 & \underline{.524} & \underline{.953} & \textbf{.210} & \underline{.855} & \underline{.300} & .869 \\
         % \cline{2-8}
         \rowcolor{verylightgray}
         & $\textbf{TolerantECG}_{\textbf{(ours)}}$ & \textbf{.813} & \textbf{.925} & \textbf{.493} & \textbf{.922} & \textbf{.348} & \textbf{.930} & \textbf{.556} & \textbf{.957} & \underline{.203} & \textbf{.865} & \textbf{.327} & \textbf{.912} \\
         \Xhline{4\arrayrulewidth}

         \multirow{7}{*}{\parbox{1.2cm}{\centering Lead-missing \& Noisy}} 
         & SimCLR & .673 & .857 & .309 & .853 & .189 & .862 & .483 & .946 & .121 & .769 & .212 & .847 \\
         % \cline{2-8}
         & BYOL & .684 & .860 & .347 & .880 & .223 & .877 & .476 & .945 & .139 & .773 & .239 & \underline{.866} \\
         % \cline{2-8}
         & CPC & .679 & .855 & .245 & .788 & .090 & .719 & .301 & .860 & .111 & .739 & .134 & .786 \\
         % \cline{2-8}
         & METS & .704 & .866 & .312 & .824 & .179 & .797 & .319 & .846 & .092 & .713 & .150 & .771 \\
         % \cline{2-8}
         & MERL & .663 & .841 & .240 & .776 & .126 & .761 & .221 & .811 & .069 & .671 & .118 & .730 \\
         % \cline{2-8}
         & ECG-FM & \underline{.744} & \underline{.892} & \textbf{.436} & \textbf{.892} & \underline{.298} & \underline{.887} & \textbf{.527} & \underline{.951} & \textbf{.182} & \textbf{.845} & \underline{.282} & .861 \\
         % \cline{2-8}
         \rowcolor{verylightgray}
         & $\textbf{TolerantECG}_{\textbf{(ours)}}$ & \textbf{.761} & \textbf{.898} & \underline{.425} & \underline{.891} & \textbf{.308} & \textbf{.911} & \underline{.508} & \textbf{.953} & \underline{.170} & \underline{.841} & \textbf{.297} & \textbf{.893} \\
         \hline
    \end{tabular}}
    
    \label{tab: ap auroc compare result}
    % \vspace{-0.2in}
\end{table*}

\begin{table}[ht]
    \caption{\textbf{AP} and \textbf{AUC} score on MIT-BIH dataset. 
    The \textbf{bold} values represent the best score, while the \underline{underlined} values denote the second best.
    }
    \centering\scalebox{0.9}{
    \begin{tabular}{c | c | c c}
        \hline
        \multirow{2}{*}{\textbf{Augmentation}} & 
         \multirow{2}{*}{\textbf{Model}} & 
         \multicolumn{2}{c}{\textbf{MIT-BIH}} \\
         % \cline{3-14}
         & & AP & AUC \\
         \Xhline{4\arrayrulewidth}

        \multirow{7}{*}{Original (2-lead)} 
         & SimCLR & .930 & .974 \\
         % \cline{2-8}
         & BYOL & .943 & .979 \\
         % \cline{2-8}
         & CPC & .827 & .934	\\
         % \cline{2-8}
         & METS & .926 & .972 \\
         % \cline{2-8}
         & MERL & .846 & .926 \\
         % \cline{2-8}
         & ECG-FM & \underline{.969} & \underline{.991} \\
         % \cline{2-8}
         \rowcolor{verylightgray}
         & $\textbf{TolerantECG}_{\textbf{(ours)}}$ & \textbf{.979} & \textbf{.994} \\
         \Xhline{4\arrayrulewidth}

         \multirow{7}{*}{Original (2-lead) \& Noisy} 
         & SimCLR & .922 & .971 \\
         % \cline{2-8}
         & BYOL & .930 & .972 \\
         % \cline{2-8}
         & CPC & .744 & .914	\\
         % \cline{2-8}
         & METS & .918 & .967 \\
         % \cline{2-8}
         & MERL & .732 & .893 \\
         % \cline{2-8}
         & ECG-FM & \underline{.956} & \underline{.984} \\
         % \cline{2-8}
         \rowcolor{verylightgray}
         & $\textbf{TolerantECG}_{\textbf{(ours)}}$ & \textbf{.966} & \textbf{.989} \\
         \hline
    \end{tabular}
    }
    \label{tab: ap auroc mit-bih}
    % \vspace{-0.2in}
\end{table}

We benchmark TolerantECG against other self-supervised learning approaches, specifically SimCLR \cite{simclr}, BYOL \cite{byol}, and CPC \cite{cpc}, using the implementation provided by \citet{self-supervised_ecg}. To ensure a fair comparison, we re-train these methods using the MIMIC-IV-ECG dataset \cite{mimic}. For SimCLR and BYOL, we replace their original encoders with ConvNeXt V2 to match the architecture used in our proposed approach. Moreover, because these approaches require two different augmentations, we utilize the original signal and alternatively contrast it with lead-missing and noisy versions.

In addition, we include comparisons with other ECG foundation models, namely METS \cite{frozen-llm}, MERL \cite{merl} and ECG-FM \cite{ecg-fm}. For MERL and ECG-FM, we use the pretrained checkpoints provided by their respective authors, whereas METS is reproduced, as it was not originally trained on the MIMIC dataset. Since ECG-FM was specifically designed to handle lead-missing condition, it serves as a strong baseline and a competitive model against TolerantECG in such scenario. 
% Although other ECG foundation models exist, such as \cite{esi}, they do not come with pre-trained weights. 

We finetune TolerantECG and the baseline models on our dataset across six different class levels of the PTB-XL dataset. The PTB-XL dataset contains 71 classes but these classes can be aggregated into other class levels, specifically \textit{Super-Diag} (5 coarse diagnostic superclasses), \textit{Sub-Diag} (23 diagnostic subclasses), \textit{Diag} (44 diagnostic labels), \textit{Rhythm} (12 rhythm statements), \textit{Form} (19 form statements), and \textit{All} (71 different SCP statements). We benchmark across all class levels of the PTB-XL dataset, offering a comprehensive performance analysis that assesses the transferability of pretrained models to diverse downstream tasks.

We also fine-tune our pretrained model on the MIT-BIH Arrhythmia Database. Since this dataset includes only two leads, it serves as a strong testbed to demonstrate our model’s tolerance to lead-missing scenarios.

% Additionally, we perform finetuning on CPSC2018 dataset. It consists of 9 classes, namely Normal, AF, I-AVB, LBBB, RBBB, PAC, PVC, STD and STE. By benchmarking on this dataset beside PTB-XL, we further evaluate the generalization capability of our approach on a different data distribution.

Tables \ref{tab: ap auroc compare result} and \ref{tab: ap auroc mit-bih} compare TolerantECG with existing baseline models under various ECG signal conditions for different downstream classification tasks in terms of AP (Area Under the Precision-Recall Curve) and AUC (Area Under the Receiver Operating Characteristic curve). 

In terms of the PTB-XL dataset, for AUC, TolerantECG outperforms all baselines across 5 out of 6 tasks and gets the second-best performance in Rhythm for \textit{Original} set. For \textit{Lead-missing} set, it takes turns with ECG-FM, a much bigger, more complicated model with the transformer architecture, to establish SOTA. Notably, TolerantECG outperforms all baselines on all tasks for the \textit{Noisy} set. Under \textit{Lead-missing \& Noisy} condition, TolerantECG achieves the best performance on 4 out of 6 tasks and is second in \textit{Sub-Diag} and \textit{Form}.

For the AP score, TolerantECG performs best in 4 tasks and is second-best for the other 2 tasks, under ECG-FM for the \textit{Original} set. However, the statement gets reversed for the \textit{Lead-missing} set. For the \textit{Noisy} set, TolerantECG performs best in 5 of 6 tasks and second-best in task \textit{Form}. For \textit{Lead-missing \& Noisy} set, TolerantECG and ECG-FM are the top-2 performing models across half of the tasks, taking turns being the best performing one.

On the MIT-BIH Arrhythmia Database, TolerantECG achieves the best performance across both augmentation settings, outperforming other methods in terms of both AP and AUC, even under the missing-lead condition. Notably, ECG-FM ranks second in both metrics. By evaluating on a dataset distinct from PTB-XL, these results further highlight the robustness of TolerantECG against lead-missing and noisy ECG conditions.

\subsubsection{Ablation Study}

To further analyze the performance of all models under varying levels of lead-missing and noise corruption, we conduct an ablation study on the dataset. This experiment evaluates each model across multiple imperfect settings by varying the number of visible leads and the Signal-to-Noise Ratio (SNR) levels. 
% The results for the \textit{All} classification task on the PTB-XL dataset are presented in Figures~\ref{fig: lead num} and~\ref{fig: snr db}, respectively.

\begin{figure}[ht]
    %\vskip 0.2in
    \begin{center}
        \centerline{\includegraphics[width=0.9\columnwidth]{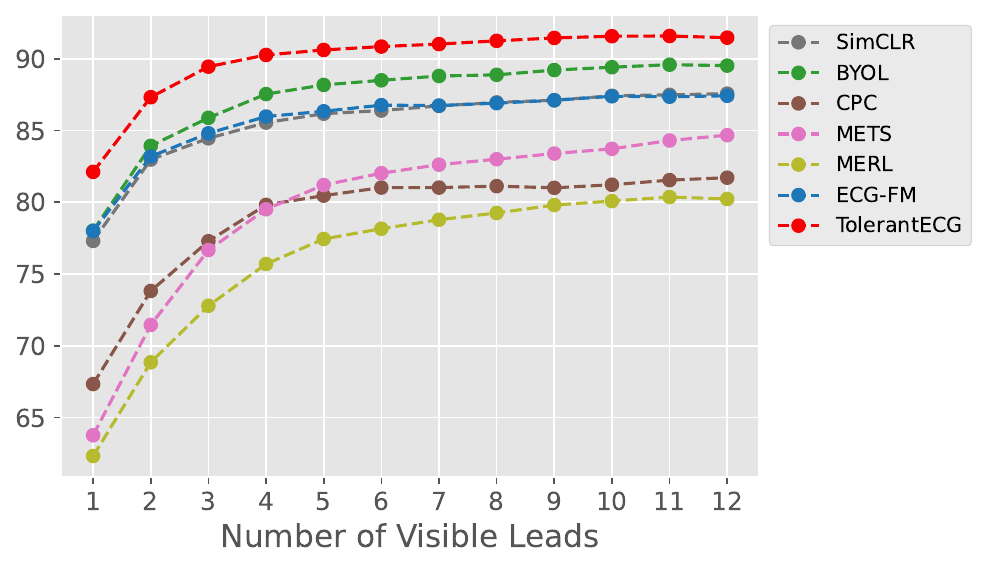}}
        \caption{AUC (\%) evaluated on different number of leads for PTB-XL All classification task}
        \label{fig: lead num}
    \end{center}
    \vspace{-0.5cm}  
\end{figure}

\begin{figure}[ht]
    %\vskip 0.2in
    \begin{center}
        \centerline{\includegraphics[width=0.9\columnwidth]{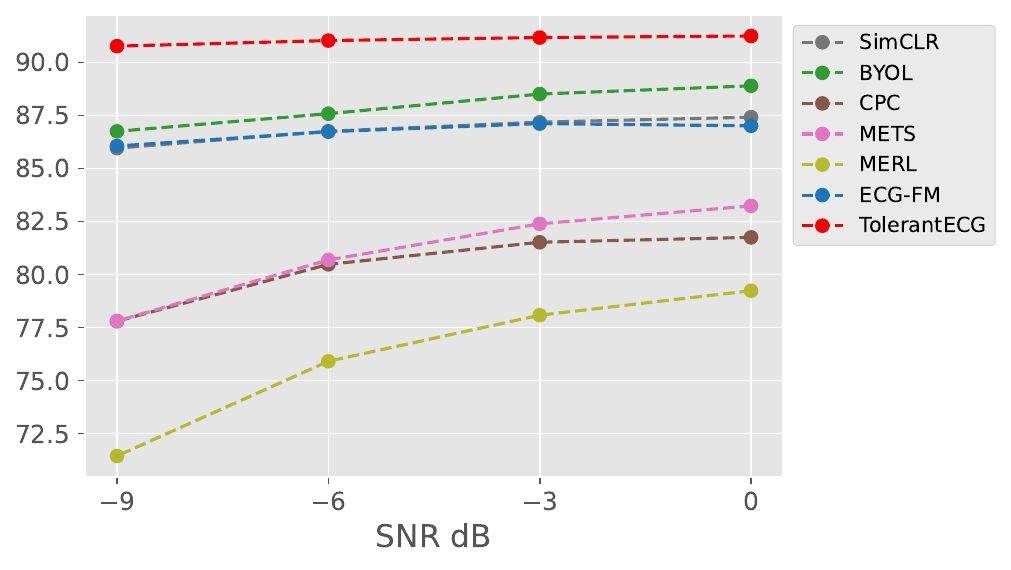}}
        \caption{AUC (\%) performance across different SNR levels for PTB-XL All classification task}
        \label{fig: snr db}
    \end{center}
    \vspace{-0.8cm}
\end{figure}

First of all, in terms of lead-missing condition shown in Figure \ref{fig: lead num} for PTB-XL All classification task, as expected, all models show improved performance as more leads become available. Notably, TolerantECG consistently outperforms all other methods across all lead configurations, especially in low-lead settings (e.g., 1–4 leads), demonstrating its strong robustness to lead-missing scenarios. ECG-FM and other methods follow behind, while SimCLR and BYOL show competitive results in higher-lead conditions. These results highlight the superior generalization capability of TolerantECG under varying signal completeness.

Secondly, in Figure \ref{fig: snr db}, as noise increases (lower SNR), all models experience performance degradation. However, TolerantECG consistently and stably maintains the highest AUC across all noise levels, showing remarkable robustness against signal corruption. BYOL, SimCLR and ECG-FM also demonstrate good tolerance to noise, while other methods like METS, CPC and MERL suffer more noticeably under severe noise. These findings underscore TolerantECG’s effectiveness in handling noisy ECG conditions.

% In summary, TolerantECG either achieves the best or second-best performance compared to all baselines across all PTB-XL downstream tasks and signal conditions. Despite its lightweight architecture—comprising only CNN layers, simple linear layers, and normalization layers—TolerantECG delivers competitive performance comparable to the significantly larger ECG-FM foundation model, which employs a more complex and powerful transformer-based architecture. This highlights the effectiveness of our training framework.

% Table~\ref{tab:params} provides a comparison of parameter counts between TolerantECG and baseline models during pretraining. Notably, during downstream transfer learning, only the ECG encoder is retained. 

% \begin{table}[htb]
%     \caption{Comparison of total parameters among models}

%     \centering
%     \begin{tabular}{|c|c|}
%         \hline
%          \textbf{Model} & \textbf{Number of parameters} \\
%          \hline
%          SimCLR & 0.89M (xresnet1d50) \\
%          \hline 
%          BYOL & 0.89M (xresnet1d50) \\
%          \hline
%         CPC & 5.00M \\
%          \hline
%          \multirow{2}{*}{MERL}
%           & 3.8M (Resnet18) \\
%           & + 110M (Med-CPT) \\
%          \hline
%          ECG-FM & 90.37M (Wav2Vec 2.0) \\
%          \hline
%          \multirow{2}{*}{$\text{TolerantECG}_{\text{(ours)}}$}
%           & 27.40M (ConvNeXt V2) \\
%           & + 108.82M (BioLinkBERT) \\
%         \hline
%     \end{tabular}
%     \label{tab:params}
%     \vspace{-0.2in}
% \end{table}

We also conduct an ablation study on ReportAlign, DuoDistill, and TolerantECG by fine-tuning each model. To evaluate the impact of CFR, we compare the pretrained ReportAlign model using raw text inputs with its CFR-augmented counterpart. We also include UniDistill in our study, which is structurally identical to DuoDistill but utilizes a single teacher model for both imperfect scenarios. 
% For this setup, a linear classification head is attached onto the ECG encoder. Finetuning updates both the self-supervised ECG encoder and the classifier head, allowing the entire model to adapt to the target task. Together, these methods provide a comprehensive analysis of how well the learned representations capture meaningful ECG features.

\begin{figure}[ht]
    %\vskip 0.2in
    \begin{center}
        \centerline{\includegraphics[width=0.9\columnwidth]{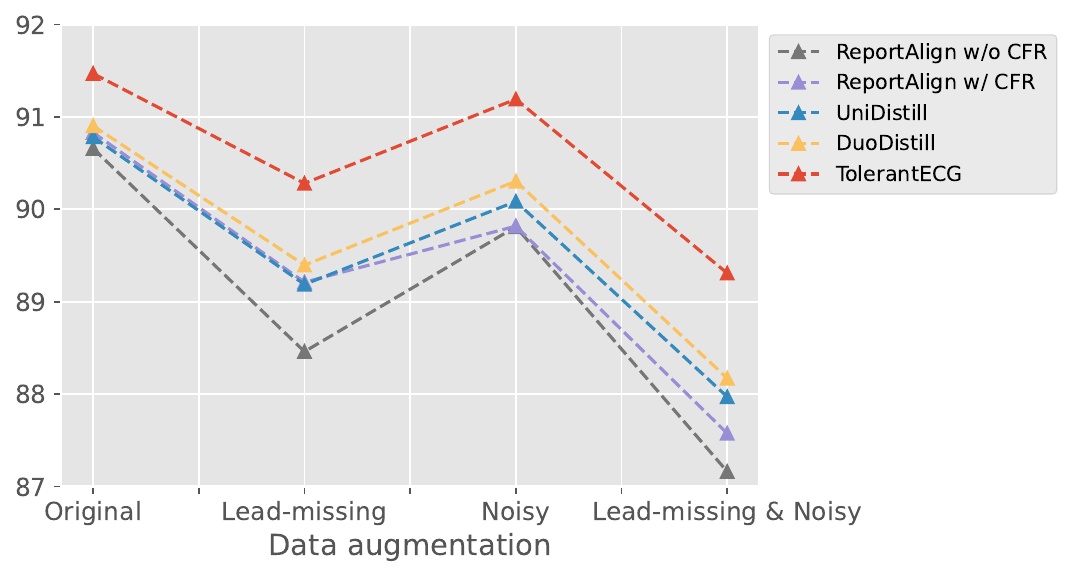}}
        \caption{\textbf{AUC} (\%) across different ECG signal conditions and training frameworks on PTB-XL All classification task}
        \label{fig: ablation auc}        
    \end{center}
    \vspace{-0.3in}
\end{figure}

The results presented as AUC scores in Figure \ref{fig: ablation auc} evaluate the performance on the All classification task of PTB-XL. As shown in the chart, TolerantECG consistently outperforms all other models under every condition, demonstrating its robustness to signal imperfections. Notably, while most models experience a substantial drop in performance under the \textit{Lead-missing \& Noisy} scenario, TolerantECG maintains relatively high AUC, indicating strong generalization under harsh corruptions. DuoDistill performs second best overall, especially under the Noisy condition, thanks to its dual-teacher architecture. UniDistill, which relies on a single teacher, generally trails behind DuoDistill and TolerantECG, suggesting that the alternating duo distillation strategy contributes meaningfully to robustness. The ReportAlign with CFR model shows a noticeable improvement over without CFR, particularly under the \textit{Lead-missing} setting, highlighting the effectiveness of CFR in enhancing alignment. However, both variants of ReportAlign degrade more significantly under combined corruptions. 

% The ablation study results, presented as AUC scores in Figure \ref{fig: ablation auc}, evaluate performance on the All classification task of PTB-XL. As shown in the chart, all models perform similarly on the original signals, with TolerantECG showing a slight improvement. However, the performance gap becomes more evident under imperfect ECG conditions. With the support of CFR, ReportAlign achieves results comparable to DuoDistill on the \textit{Lead-missing} test set. Notably, ReportAlign experiences performance degradation under imperfect conditions, whereas DuoDistill remains more stable. By combining these two approaches, TolerantECG leverages their strengths, achieving both a high-quality understanding of ECG signals and robustness against lead-missing and noisy scenarios.

\section{Conclusion}
This paper introduces TolerantECG, a novel training approach for an ECG foundation model capable of learning robust ECG representations from noisy and/or lead-missing ECG recordings. The method combines contrastive learning guided by detailed cardiac text reports and self-supervised learning with dual-mode distillation that handles the two types of imperfect ECG conditions. Extensive experiments provide a comprehensive analysis of the model’s downstream transferability across all class levels (i.e. downstream tasks) of the PTB-XL dataset. Results demonstrate that the proposed method consistently ranks as the best or second-best performer across all the downstream tasks under various ECG signal conditions. An ablation study highlights that the dual-mode distillation module contributes significantly more to overall performance than the report-alignment contrastive learning module in the TolerantECG framework.

For future work, we plan to enhance the TolerantECG training framework by using a transformer-based architecture for the ECG encoder to explore potential performance improvements.

%
% broader impact: promising to solve out of distribution
%\section*{Broader Impacts}
%By analyzing ECG signals in real-time, it can identify potential health risks at an early stage, allowing for timely intervention and reducing the risk of severe heart conditions.

\section*{Limitations} 
Our CFR module currently depends on a third-party ECG diagnosis database, which may have limitations in clinical comprehensiveness. However, the CFR can immediately benefit from enhancements to this database, whether through the inclusion of additional data or by integrating a more comprehensive ECG diagnosis database. In addition, the training database for TolerantECG currently includes only three noise types from the MIT-BIH noise dataset. Nevertheless, our framework is designed to be easily extensible, allowing for the integration of additional noise types as needed.

\bibliographystyle{ACM-Reference-Format}
\balance
\bibliography{ref}


\begin{thebibliography}{35}

%%% the Web address.

\ifx \showCODEN    \undefined \def \showCODEN     #1{\unskip}     \fi
\ifx \showISBNx    \undefined \def \showISBNx     #1{\unskip}     \fi
\ifx \showISBNxiii \undefined \def \showISBNxiii  #1{\unskip}     \fi
\ifx \showISSN     \undefined \def \showISSN      #1{\unskip}     \fi
\ifx \showLCCN     \undefined \def \showLCCN      #1{\unskip}     \fi
\ifx \shownote     \undefined \def \shownote      #1{#1}          \fi
\ifx \showarticletitle \undefined \def \showarticletitle #1{#1}   \fi
\ifx \showURL      \undefined \def \showURL       {\relax}        \fi
% invisible to TeX
\providecommand\bibfield[2]{#2}
\providecommand\bibinfo[2]{#2}
\providecommand\natexlab[1]{#1}
\providecommand\showeprint[2][]{arXiv:#2}

\bibitem[LIT({[n.\,d.]})]%
        {LITFL}
 \bibinfo{year}{[n.\,d.]}\natexlab{}.
\newblock \bibinfo{title}{{Life in the Fastlane}}.
\newblock \bibinfo{howpublished}{\url{https://litfl.com/ecg-library/}}.
\newblock
\newblock
\shownote{Accessed: 2024-10-15}.


\bibitem[Alickovic and Babic(2015)]%
        {noise-impact2}
\bibfield{author}{\bibinfo{person}{Emina Alickovic} {and} \bibinfo{person}{Zdenka Babic}.} \bibinfo{year}{2015}\natexlab{}.
\newblock \showarticletitle{The effect of denoising on classification of ECG signals}. In \bibinfo{booktitle}{\emph{2015 XXV International Conference on Information, Communication and Automation Technologies (ICAT)}}. \bibinfo{pages}{1--6}.
\newblock
\href{https://doi.org/10.1109/ICAT.2015.7340540}{doi:\nolinkurl{10.1109/ICAT.2015.7340540}}


\bibitem[Cappellani et~al\mbox{.}(2024)]%
        {Cappellani2024}
\bibfield{author}{\bibinfo{person}{Francesco Cappellani}, \bibinfo{person}{Kevin~R. Card}, \bibinfo{person}{Carol~L. Shields}, \bibinfo{person}{Jose~S. Pulido}, {and} \bibinfo{person}{Julia~A. Haller}.} \bibinfo{year}{2024}\natexlab{}.
\newblock \showarticletitle{Reliability and accuracy of artificial intelligence ChatGPT in providing information on ophthalmic diseases and management to patients}.
\newblock \bibinfo{journal}{\emph{Eye}} \bibinfo{volume}{38}, \bibinfo{number}{7} (\bibinfo{year}{2024}), \bibinfo{pages}{1368--1373}.
\newblock
\showISSN{1476-5454}
\href{https://doi.org/10.1038/s41433-023-02906-0}{doi:\nolinkurl{10.1038/s41433-023-02906-0}}


\bibitem[Caron et~al\mbox{.}(2021)]%
        {dino}
\bibfield{author}{\bibinfo{person}{Mathilde Caron}, \bibinfo{person}{Hugo Touvron}, \bibinfo{person}{Ishan Misra}, \bibinfo{person}{Hervé Jégou}, \bibinfo{person}{Julien Mairal}, \bibinfo{person}{Piotr Bojanowski}, {and} \bibinfo{person}{Armand Joulin}.} \bibinfo{year}{2021}\natexlab{}.
\newblock \bibinfo{title}{Emerging Properties in Self-Supervised Vision Transformers}.
\newblock
\showeprint[arxiv]{2104.14294}~[cs.CV]
\urldef\tempurl%
\url{https://arxiv.org/abs/2104.14294}
\showURL{%
\tempurl}


\bibitem[Chen et~al\mbox{.}(2020)]%
        {simclr}
\bibfield{author}{\bibinfo{person}{Ting Chen}, \bibinfo{person}{Simon Kornblith}, \bibinfo{person}{Mohammad Norouzi}, {and} \bibinfo{person}{Geoffrey Hinton}.} \bibinfo{year}{2020}\natexlab{}.
\newblock \bibinfo{title}{A Simple Framework for Contrastive Learning of Visual Representations}.
\newblock
\showeprint[arxiv]{2002.05709}~[cs.LG]
\urldef\tempurl%
\url{https://arxiv.org/abs/2002.05709}
\showURL{%
\tempurl}


\bibitem[Clifford et~al\mbox{.}(2006)]%
        {ecg-noises}
\bibfield{author}{\bibinfo{person}{Gari~D Clifford}, \bibinfo{person}{Francisco Azuaje}, \bibinfo{person}{Patrick Mcsharry}, {et~al\mbox{.}}} \bibinfo{year}{2006}\natexlab{}.
\newblock \showarticletitle{ECG statistics, noise, artifacts, and missing data}.
\newblock \bibinfo{journal}{\emph{Advanced methods and tools for ECG data analysis}} \bibinfo{volume}{6}, \bibinfo{number}{1} (\bibinfo{year}{2006}), \bibinfo{pages}{18}.
\newblock


\bibitem[Devlin et~al\mbox{.}(2019)]%
        {bert}
\bibfield{author}{\bibinfo{person}{Jacob Devlin}, \bibinfo{person}{Ming-Wei Chang}, \bibinfo{person}{Kenton Lee}, {and} \bibinfo{person}{Kristina Toutanova}.} \bibinfo{year}{2019}\natexlab{}.
\newblock \bibinfo{title}{BERT: Pre-training of Deep Bidirectional Transformers for Language Understanding}.
\newblock
\showeprint[arxiv]{1810.04805}~[cs.CL]
\urldef\tempurl%
\url{https://arxiv.org/abs/1810.04805}
\showURL{%
\tempurl}


\bibitem[Farrell and Rowlandson(2006)]%
        {noise-impact3}
\bibfield{author}{\bibinfo{person}{Robert~M. Farrell} {and} \bibinfo{person}{G.~Ian Rowlandson}.} \bibinfo{year}{2006}\natexlab{}.
\newblock \showarticletitle{The effects of noise on computerized electrocardiogram measurements}.
\newblock \bibinfo{journal}{\emph{Journal of Electrocardiology}} \bibinfo{volume}{39}, \bibinfo{number}{4, Supplement} (\bibinfo{year}{2006}), \bibinfo{pages}{S165--S173}.
\newblock
\showISSN{0022-0736}
\href{https://doi.org/10.1016/j.jelectrocard.2006.05.025}{doi:\nolinkurl{10.1016/j.jelectrocard.2006.05.025}}
\newblock
\shownote{Research and Technology Transfer in Computerized Electrocardiology}.


\bibitem[Goldberger et~al\mbox{.}(2000)]%
        {physionet}
\bibfield{author}{\bibinfo{person}{A. Goldberger}, \bibinfo{person}{L. Amaral}, \bibinfo{person}{L. Glass}, \bibinfo{person}{J. Hausdorff}, \bibinfo{person}{P.~C. Ivanov}, \bibinfo{person}{R. Mark}, {and} \bibinfo{person}{H.~E. Stanley}.} \bibinfo{year}{2000}\natexlab{}.
\newblock \showarticletitle{PhysioBank, PhysioToolkit, and PhysioNet: Components of a new research resource for complex physiologic signals}.
\newblock \bibinfo{journal}{\emph{Circulation}} \bibinfo{volume}{101}, \bibinfo{number}{23} (\bibinfo{year}{2000}), \bibinfo{pages}{e215--e220}.
\newblock
\newblock
\shownote{Online}.


\bibitem[Gow et~al\mbox{.}(2023)]%
        {mimic}
\bibfield{author}{\bibinfo{person}{Brian Gow}, \bibinfo{person}{Tom Pollard}, \bibinfo{person}{Larry~A Nathanson}, \bibinfo{person}{Alistair Johnson}, \bibinfo{person}{Benjamin Moody}, \bibinfo{person}{Chrystinne Fernandes}, \bibinfo{person}{Nathaniel Greenbaum}, \bibinfo{person}{Seth Berkowitz}, \bibinfo{person}{Dana Moukheiber}, \bibinfo{person}{Parastou Eslami}, {et~al\mbox{.}}} \bibinfo{year}{2023}\natexlab{}.
\newblock \showarticletitle{Mimic-iv-ecg-diagnostic electrocardiogram matched subset}.
\newblock \bibinfo{journal}{\emph{Type: dataset}} (\bibinfo{year}{2023}).
\newblock


\bibitem[Grill et~al\mbox{.}(2020)]%
        {byol}
\bibfield{author}{\bibinfo{person}{Jean-Bastien Grill}, \bibinfo{person}{Florian Strub}, \bibinfo{person}{Florent Altché}, \bibinfo{person}{Corentin Tallec}, \bibinfo{person}{Pierre~H. Richemond}, \bibinfo{person}{Elena Buchatskaya}, \bibinfo{person}{Carl Doersch}, \bibinfo{person}{Bernardo~Avila Pires}, \bibinfo{person}{Zhaohan~Daniel Guo}, \bibinfo{person}{Mohammad~Gheshlaghi Azar}, \bibinfo{person}{Bilal Piot}, \bibinfo{person}{Koray Kavukcuoglu}, \bibinfo{person}{Rémi Munos}, {and} \bibinfo{person}{Michal Valko}.} \bibinfo{year}{2020}\natexlab{}.
\newblock \bibinfo{title}{Bootstrap your own latent: A new approach to self-supervised Learning}.
\newblock
\showeprint[arxiv]{2006.07733}~[cs.LG]
\urldef\tempurl%
\url{https://arxiv.org/abs/2006.07733}
\showURL{%
\tempurl}


\bibitem[Li et~al\mbox{.}(2023)]%
        {frozen-llm}
\bibfield{author}{\bibinfo{person}{Jun Li}, \bibinfo{person}{Che Liu}, \bibinfo{person}{Sibo Cheng}, \bibinfo{person}{Rossella Arcucci}, {and} \bibinfo{person}{Shenda Hong}.} \bibinfo{year}{2023}\natexlab{}.
\newblock \bibinfo{title}{Frozen Language Model Helps ECG Zero-Shot Learning}.
\newblock
\showeprint[arxiv]{2303.12311}~[cs.LG]
\urldef\tempurl%
\url{https://arxiv.org/abs/2303.12311}
\showURL{%
\tempurl}


\bibitem[Lin et~al\mbox{.}(2024)]%
        {denoise-mamba}
\bibfield{author}{\bibinfo{person}{Jie Lin}, \bibinfo{person}{I Chiu}, \bibinfo{person}{Kuan-Chen Wang}, \bibinfo{person}{Kai-Chun Liu}, \bibinfo{person}{Hsin-Min Wang}, \bibinfo{person}{Ping-Cheng Yeh}, {and} \bibinfo{person}{Yu Tsao}.} \bibinfo{year}{2024}\natexlab{}.
\newblock \bibinfo{title}{MSECG: Incorporating Mamba for Robust and Efficient ECG Super-Resolution}.
\newblock
\showeprint[arxiv]{2412.04861}~[cs.LG]
\urldef\tempurl%
\url{https://arxiv.org/abs/2412.04861}
\showURL{%
\tempurl}


\bibitem[Liu et~al\mbox{.}(2024)]%
        {merl}
\bibfield{author}{\bibinfo{person}{Che Liu}, \bibinfo{person}{Zhongwei Wan}, \bibinfo{person}{Cheng Ouyang}, \bibinfo{person}{Anand Shah}, \bibinfo{person}{Wenjia Bai}, {and} \bibinfo{person}{Rossella Arcucci}.} \bibinfo{year}{2024}\natexlab{}.
\newblock \bibinfo{title}{Zero-Shot ECG Classification with Multimodal Learning and Test-time Clinical Knowledge Enhancement}.
\newblock
\showeprint[arxiv]{2403.06659}~[eess.SP]
\urldef\tempurl%
\url{https://arxiv.org/abs/2403.06659}
\showURL{%
\tempurl}


\bibitem[McKeen et~al\mbox{.}(2024)]%
        {ecg-fm}
\bibfield{author}{\bibinfo{person}{Kaden McKeen}, \bibinfo{person}{Laura Oliva}, \bibinfo{person}{Sameer Masood}, \bibinfo{person}{Augustin Toma}, \bibinfo{person}{Barry Rubin}, {and} \bibinfo{person}{Bo Wang}.} \bibinfo{year}{2024}\natexlab{}.
\newblock \bibinfo{title}{ECG-FM: An Open Electrocardiogram Foundation Model}.
\newblock
\showeprint[arxiv]{2408.05178}~[cs.LG]
\urldef\tempurl%
\url{https://arxiv.org/abs/2408.05178}
\showURL{%
\tempurl}


\bibitem[McKenna et~al\mbox{.}(2024)]%
        {noise-impact}
\bibfield{author}{\bibinfo{person}{Stacey McKenna}, \bibinfo{person}{Naomi McCord}, \bibinfo{person}{Jordan Diven}, \bibinfo{person}{Matthew Fitzpatrick}, \bibinfo{person}{Holly Easlea}, \bibinfo{person}{Austin Gibbs}, {and} \bibinfo{person}{Andrew R~J Mitchell}.} \bibinfo{year}{2024}\natexlab{}.
\newblock \showarticletitle{Evaluating the impacts of digital ECG denoising on the interpretive capabilities of healthcare professionals}.
\newblock \bibinfo{journal}{\emph{European Heart Journal - Digital Health}} \bibinfo{volume}{5}, \bibinfo{number}{5} (\bibinfo{date}{08} \bibinfo{year}{2024}), \bibinfo{pages}{601--610}.
\newblock
\showISSN{2634-3916}
\href{https://doi.org/10.1093/ehjdh/ztae063}{doi:\nolinkurl{10.1093/ehjdh/ztae063}}
\showeprint{https://academic.oup.com/ehjdh/article-pdf/5/5/601/59236875/ztae063.pdf}


\bibitem[Mehari and Strodthoff(2022)]%
        {self-supervised_ecg}
\bibfield{author}{\bibinfo{person}{Temesgen Mehari} {and} \bibinfo{person}{Nils Strodthoff}.} \bibinfo{year}{2022}\natexlab{}.
\newblock \showarticletitle{Self-supervised representation learning from 12-lead ECG data}.
\newblock \bibinfo{journal}{\emph{Computers in Biology and Medicine}}  \bibinfo{volume}{141} (\bibinfo{date}{Feb.} \bibinfo{year}{2022}), \bibinfo{pages}{105114}.
\newblock
\showISSN{0010-4825}
\href{https://doi.org/10.1016/j.compbiomed.2021.105114}{doi:\nolinkurl{10.1016/j.compbiomed.2021.105114}}


\bibitem[Mohamoud et~al\mbox{.}(2024)]%
        {smartwatch}
\bibfield{author}{\bibinfo{person}{Abdilahi Mohamoud}, \bibinfo{person}{Joseph Jensen}, {and} \bibinfo{person}{Kevin~G. Buda}.} \bibinfo{year}{2024}\natexlab{}.
\newblock \showarticletitle{Consumer-grade wearable cardiac monitors: What they do well, and what needs work}.
\newblock \bibinfo{journal}{\emph{Cleveland Clinic Journal of Medicine}} \bibinfo{volume}{91}, \bibinfo{number}{1} (\bibinfo{year}{2024}), \bibinfo{pages}{23--29}.
\newblock
\showISSN{0891-1150}
\href{https://doi.org/10.3949/ccjm.91a.23030}{doi:\nolinkurl{10.3949/ccjm.91a.23030}}
\showeprint{https://www.ccjm.org/content/91/1/23.full.pdf}


\bibitem[Moody and Mark(2001)]%
        {mit-bih-data}
\bibfield{author}{\bibinfo{person}{G.B. Moody} {and} \bibinfo{person}{R.G. Mark}.} \bibinfo{year}{2001}\natexlab{}.
\newblock \showarticletitle{The impact of the MIT-BIH Arrhythmia Database}.
\newblock \bibinfo{journal}{\emph{IEEE Engineering in Medicine and Biology Magazine}} \bibinfo{volume}{20}, \bibinfo{number}{3} (\bibinfo{year}{2001}), \bibinfo{pages}{45--50}.
\newblock
\href{https://doi.org/10.1109/51.932724}{doi:\nolinkurl{10.1109/51.932724}}


\bibitem[Moody et~al\mbox{.}(1984)]%
        {mit-bih}
\bibfield{author}{\bibinfo{person}{GB Moody}, \bibinfo{person}{WE Muldrow}, {and} \bibinfo{person}{RG Mark}.} \bibinfo{year}{1984}\natexlab{}.
\newblock \showarticletitle{A noise stress test for arrhythmia detectors}.
\newblock \bibinfo{journal}{\emph{Computers in Cardiology}}  \bibinfo{volume}{11} (\bibinfo{year}{1984}), \bibinfo{pages}{381--384}.
\newblock
\urldef\tempurl%
\url{http://ecg.mit.edu/george/publications/nst-cinc-1984.pdf}
\showURL{%
\tempurl}


\bibitem[Nazarian et~al\mbox{.}(2021)]%
        {smartwatch2}
\bibfield{author}{\bibinfo{person}{Scarlet Nazarian}, \bibinfo{person}{Kyle Lam}, \bibinfo{person}{Ara Darzi}, {and} \bibinfo{person}{Hutan Ashrafian}.} \bibinfo{year}{2021}\natexlab{}.
\newblock \showarticletitle{Diagnostic Accuracy of Smartwatches for the Detection of Cardiac Arrhythmia: Systematic Review and Meta-analysis}.
\newblock \bibinfo{journal}{\emph{J Med Internet Res}} \bibinfo{volume}{23}, \bibinfo{number}{8} (\bibinfo{date}{27 Aug} \bibinfo{year}{2021}), \bibinfo{pages}{e28974}.
\newblock
\showISSN{1438-8871}
\href{https://doi.org/10.2196/28974}{doi:\nolinkurl{10.2196/28974}}


\bibitem[Radford et~al\mbox{.}(2021)]%
        {clip}
\bibfield{author}{\bibinfo{person}{Alec Radford}, \bibinfo{person}{Jong~Wook Kim}, \bibinfo{person}{Chris Hallacy}, \bibinfo{person}{Aditya Ramesh}, \bibinfo{person}{Gabriel Goh}, \bibinfo{person}{Sandhini Agarwal}, \bibinfo{person}{Girish Sastry}, \bibinfo{person}{Amanda Askell}, \bibinfo{person}{Pamela Mishkin}, \bibinfo{person}{Jack Clark}, \bibinfo{person}{Gretchen Krueger}, {and} \bibinfo{person}{Ilya Sutskever}.} \bibinfo{year}{2021}\natexlab{}.
\newblock \bibinfo{title}{Learning Transferable Visual Models From Natural Language Supervision}.
\newblock
\showeprint[arxiv]{2103.00020}~[cs.CV]
\urldef\tempurl%
\url{https://arxiv.org/abs/2103.00020}
\showURL{%
\tempurl}


\bibitem[Reimers and Gurevych(2019)]%
        {sentence-bert}
\bibfield{author}{\bibinfo{person}{Nils Reimers} {and} \bibinfo{person}{Iryna Gurevych}.} \bibinfo{year}{2019}\natexlab{}.
\newblock \showarticletitle{Sentence-BERT: Sentence Embeddings using Siamese BERT-Networks}. In \bibinfo{booktitle}{\emph{Proceedings of the 2019 Conference on Empirical Methods in Natural Language Processing}}. \bibinfo{publisher}{Association for Computational Linguistics}.
\newblock
\urldef\tempurl%
\url{https://arxiv.org/abs/1908.10084}
\showURL{%
\tempurl}


\bibitem[Ródenas et~al\mbox{.}(2018)]%
        {denoise-dwt}
\bibfield{author}{\bibinfo{person}{Juan Ródenas}, \bibinfo{person}{Manuel García}, \bibinfo{person}{José J~Rieta}, {and} \bibinfo{person}{Raul Alcaraz}.} \bibinfo{year}{2018}\natexlab{}.
\newblock \showarticletitle{An Efficient Algorithm Based on Wavelet Transform to Reduce Powerline Noise From Electrocardiograms}. In \bibinfo{booktitle}{\emph{2018 Computing in Cardiology Conference (CinC)}} \emph{(\bibinfo{series}{CinC2018})}. \bibinfo{publisher}{Computing in Cardiology}.
\newblock
\showISSN{2325-887X}
\href{https://doi.org/10.22489/cinc.2018.200}{doi:\nolinkurl{10.22489/cinc.2018.200}}


\bibitem[Shen et~al\mbox{.}(2023)]%
        {shen2023chatgpttrustmeasuringcharacterizing}
\bibfield{author}{\bibinfo{person}{Xinyue Shen}, \bibinfo{person}{Zeyuan Chen}, \bibinfo{person}{Michael Backes}, {and} \bibinfo{person}{Yang Zhang}.} \bibinfo{year}{2023}\natexlab{}.
\newblock \bibinfo{title}{In ChatGPT We Trust? Measuring and Characterizing the Reliability of ChatGPT}.
\newblock
\showeprint[arxiv]{2304.08979}~[cs.CR]
\urldef\tempurl%
\url{https://arxiv.org/abs/2304.08979}
\showURL{%
\tempurl}


\bibitem[Strik et~al\mbox{.}(2024)]%
        {smartwatch_beyond_afib}
\bibfield{author}{\bibinfo{person}{Marc Strik}, \bibinfo{person}{Sylvain Ploux}, \bibinfo{person}{Daniel Weigel}, \bibinfo{person}{Joske {van der Zande}}, \bibinfo{person}{Anouk Velraeds}, \bibinfo{person}{Hugo-Pierre Racine}, \bibinfo{person}{F.~Daniel Ramirez}, \bibinfo{person}{Michel Haïssaguerre}, {and} \bibinfo{person}{Pierre Bordachar}.} \bibinfo{year}{2024}\natexlab{}.
\newblock \showarticletitle{The use of smartwatch electrocardiogram beyond arrhythmia detection}.
\newblock \bibinfo{journal}{\emph{Trends in Cardiovascular Medicine}} \bibinfo{volume}{34}, \bibinfo{number}{3} (\bibinfo{year}{2024}), \bibinfo{pages}{174--180}.
\newblock
\showISSN{1050-1738}
\href{https://doi.org/10.1016/j.tcm.2022.12.006}{doi:\nolinkurl{10.1016/j.tcm.2022.12.006}}


\bibitem[van~den Oord et~al\mbox{.}(2019)]%
        {cpc}
\bibfield{author}{\bibinfo{person}{Aaron van~den Oord}, \bibinfo{person}{Yazhe Li}, {and} \bibinfo{person}{Oriol Vinyals}.} \bibinfo{year}{2019}\natexlab{}.
\newblock \bibinfo{title}{Representation Learning with Contrastive Predictive Coding}.
\newblock
\showeprint[arxiv]{1807.03748}~[cs.LG]
\urldef\tempurl%
\url{https://arxiv.org/abs/1807.03748}
\showURL{%
\tempurl}


\bibitem[Wagner et~al\mbox{.}(2022)]%
        {ptb-xl}
\bibfield{author}{\bibinfo{person}{Patrick Wagner}, \bibinfo{person}{Nils Strodthoff}, \bibinfo{person}{Ralf-Dieter Bousseljot}, \bibinfo{person}{Wojciech Samek}, {and} \bibinfo{person}{Tobias Schaeffter}.} \bibinfo{year}{2022}\natexlab{}.
\newblock \bibinfo{title}{PTB-XL, a large publicly available electrocardiography dataset (version 1.0.3)}.
\newblock
\href{https://doi.org/10.13026/kfzx-aw45}{doi:\nolinkurl{10.13026/kfzx-aw45}}


\bibitem[Witvliet et~al\mbox{.}(2021)]%
        {1-lead}
\bibfield{author}{\bibinfo{person}{M.~Patrick Witvliet}, \bibinfo{person}{Evert~P.M. Karregat}, \bibinfo{person}{Jelle~C.L. Himmelreich}, \bibinfo{person}{Jonas~S.S.G. {de Jong}}, \bibinfo{person}{Wim~A.M. Lucassen}, {and} \bibinfo{person}{Ralf~E. Harskamp}.} \bibinfo{year}{2021}\natexlab{}.
\newblock \showarticletitle{Usefulness, pitfalls and interpretation of handheld single‑lead electrocardiograms}.
\newblock \bibinfo{journal}{\emph{Journal of Electrocardiology}}  \bibinfo{volume}{66} (\bibinfo{year}{2021}), \bibinfo{pages}{33--37}.
\newblock
\showISSN{0022-0736}
\href{https://doi.org/10.1016/j.jelectrocard.2021.02.011}{doi:\nolinkurl{10.1016/j.jelectrocard.2021.02.011}}


\bibitem[Woo et~al\mbox{.}(2023)]%
        {convnextv2}
\bibfield{author}{\bibinfo{person}{Sanghyun Woo}, \bibinfo{person}{Shoubhik Debnath}, \bibinfo{person}{Ronghang Hu}, \bibinfo{person}{Xinlei Chen}, \bibinfo{person}{Zhuang Liu}, \bibinfo{person}{In~So Kweon}, {and} \bibinfo{person}{Saining Xie}.} \bibinfo{year}{2023}\natexlab{}.
\newblock \bibinfo{title}{ConvNeXt V2: Co-designing and Scaling ConvNets with Masked Autoencoders}.
\newblock
\showeprint[arxiv]{2301.00808}~[cs.CV]
\urldef\tempurl%
\url{https://arxiv.org/abs/2301.00808}
\showURL{%
\tempurl}


\bibitem[Yasunaga et~al\mbox{.}(2022)]%
        {biolinkbert}
\bibfield{author}{\bibinfo{person}{Michihiro Yasunaga}, \bibinfo{person}{Jure Leskovec}, {and} \bibinfo{person}{Percy Liang}.} \bibinfo{year}{2022}\natexlab{}.
\newblock \bibinfo{title}{LinkBERT: Pretraining Language Models with Document Links}.
\newblock
\showeprint[arxiv]{2203.15827}~[cs.CL]
\urldef\tempurl%
\url{https://arxiv.org/abs/2203.15827}
\showURL{%
\tempurl}


\bibitem[Yu et~al\mbox{.}(2023)]%
        {esi-rag}
\bibfield{author}{\bibinfo{person}{Han Yu}, \bibinfo{person}{Peikun Guo}, {and} \bibinfo{person}{Akane Sano}.} \bibinfo{year}{2023}\natexlab{}.
\newblock \showarticletitle{Zero-Shot ECG Diagnosis with Large Language Models and Retrieval-Augmented Generation}. In \bibinfo{booktitle}{\emph{Proceedings of the 3rd Machine Learning for Health Symposium}} \emph{(\bibinfo{series}{Proceedings of Machine Learning Research}, Vol.~\bibinfo{volume}{225})}, \bibfield{editor}{\bibinfo{person}{Stefan Hegselmann}, \bibinfo{person}{Antonio Parziale}, \bibinfo{person}{Divya Shanmugam}, \bibinfo{person}{Shengpu Tang}, \bibinfo{person}{Mercy~Nyamewaa Asiedu}, \bibinfo{person}{Serina Chang}, \bibinfo{person}{Tom Hartvigsen}, {and} \bibinfo{person}{Harvineet Singh}} (Eds.). \bibinfo{publisher}{PMLR}, \bibinfo{pages}{650--663}.
\newblock
\urldef\tempurl%
\url{https://proceedings.mlr.press/v225/yu23b.html}
\showURL{%
\tempurl}


\bibitem[Yu et~al\mbox{.}(2024)]%
        {esi}
\bibfield{author}{\bibinfo{person}{Han Yu}, \bibinfo{person}{Peikun Guo}, {and} \bibinfo{person}{Akane Sano}.} \bibinfo{year}{2024}\natexlab{}.
\newblock \bibinfo{title}{ECG Semantic Integrator (ESI): A Foundation ECG Model Pretrained with LLM-Enhanced Cardiological Text}.
\newblock
\showeprint[arxiv]{2405.19366}~[eess.SP]
\urldef\tempurl%
\url{https://arxiv.org/abs/2405.19366}
\showURL{%
\tempurl}


\bibitem[Zhao et~al\mbox{.}(2024)]%
        {ecg-chat}
\bibfield{author}{\bibinfo{person}{Yubao Zhao}, \bibinfo{person}{Tian Zhang}, \bibinfo{person}{Xu Wang}, \bibinfo{person}{Puyu Han}, \bibinfo{person}{Tong Chen}, \bibinfo{person}{Linlin Huang}, \bibinfo{person}{Youzhu Jin}, {and} \bibinfo{person}{Jiaju Kang}.} \bibinfo{year}{2024}\natexlab{}.
\newblock \bibinfo{title}{ECG-Chat: A Large ECG-Language Model for Cardiac Disease Diagnosis}.
\newblock
\showeprint[arxiv]{2408.08849}~[eess.SP]
\urldef\tempurl%
\url{https://arxiv.org/abs/2408.08849}
\showURL{%
\tempurl}


\bibitem[Zhu et~al\mbox{.}(2024)]%
        {denoise-transformer}
\bibfield{author}{\bibinfo{person}{Ding Zhu}, \bibinfo{person}{Vishnu~Kabir Chhabra}, {and} \bibinfo{person}{Mohammad~Mahdi Khalili}.} \bibinfo{year}{2024}\natexlab{}.
\newblock \bibinfo{title}{ECG Signal Denoising Using Multi-scale Patch Embedding and Transformers}.
\newblock
\showeprint[arxiv]{2407.11065}~[eess.SP]
\urldef\tempurl%
\url{https://arxiv.org/abs/2407.11065}
\showURL{%
\tempurl}


\end{thebibliography}

% \appendix
% \onecolumn
% \section{ECG data example} \label{ECG sample}

% \section{Results on different levels of Lead-missing and Noisy} \label{robust prove}

% To demonstrate the robustness of TolerantECG across varying augmentation levels, we evaluate all finetuned models on the \textit{Lead-missing} and \textit{Noisy} augmented PTB-XL's All-level classification task. The evaluation is conducted separately across different numbers of leads and SNR levels (dB). The results are presented in Figures \ref{fig: lead num} and \ref{fig: snr db}.

% \begin{figure*}[ht]
%     %\vskip 0.2in
%     \begin{center}
%         \centerline{\includegraphics[width=\textwidth]{fig/appendix lead.pdf}}
%         \caption{AUC (\%) evaluated on different number of leads for PTB-XL All classification task}
%         \label{fig: lead num}
%     \end{center}
%     %\vskip -0.2in    
% \end{figure*}

% \begin{figure*}[ht]
%     %\vskip 0.2in
%     \begin{center}
%         \centerline{\includegraphics[width=\textwidth]{fig/appendix snr.pdf}}
%         \caption{AUC (\%) evaluated on different SNR dB for PTB-XL All classification task}
%         \label{fig: snr db}
%     \end{center}
%     %\vskip -0.2in    
% \end{figure*}

\end{document}